Check for updates

OPEN

# A comprehensive survey of deep face verification systems adversarial attacks and defense strategies

Sohair Kilany[1]✉ & Ahmed Mahfouz[1,2]✉

**Face Verification (FV) systems have exhibited remarkable performance in verification tasks and have consequently garnered extensive adoption across various applications, from identity duplication to authentication in mobile payments. However, the surge in popularity of face verification has raised concerns about potential vulnerabilities in the face of adversarial attacks. These concerns originate from the fact that advanced FV systems, which rely on deep neural networks, have recently demonstrated susceptibility to crafted input samples known as adversarial examples. Although imperceptible to human observers, adversarial examples can deceive deep neural networks during the testing and deployment phases. These vulnerabilities raised significant concerns about the deployment of deep neural networks in safety-critical contexts, prompting extensive investigations into adversarial attacks and corresponding defense strategies. This comprehensive survey provides a comprehensive overview of recent advances in deep face verification, encompassing a broad spectrum of topics such as algorithmic designs, database utilization, protocols, and application scenarios. Furthermore, we conduct an in-depth examination of state-of-the-art algorithms to generate adversarial examples and the defense mechanisms devised to mitigate such adversarial threats.**

Face verification (FV) is an active research topic in computer vision. It has been involved in various applications such as active authentication[1–4], driving licenses, and airport security due to the growing size of face databases and the high accuracy of the FV system. Different organizations around the world have widely adopted FV.

The key to face verification is extracting a discriminative set of features from face images using Deep Convolutional Neural Networks (DCNNs). With recent advances in DCNNs[5,6], face verification reached impressive performances and highly accurate results over the past years[7–10]. For instance, the accuracy of Labeled Faces in the Wild (LFW) benchmark dataset[11] has been boosted from 97% to above 99.8%, and Youtube Faces (YTF) dataset[12] has been increased from 91.4% to above 98%. Moreover, DCNN frameworks enable end-to-end learning, i.e., learning a mapping from the input image space to the target label space.

In the last few years, researchers have discovered that FV systems are susceptible to attacks that introduce data variations, which can deceive classifiers. These attacks can be accomplished either via (i) Spoof attacks: artifacts in the physical domain (i.e., 3D masks, eyeglasses, replaying videos)[13], (ii) Adversarial perturbation attacks: imperceptible noises added to probes for evading FV systems, and (iii) Digital manipulation attacks: entirely or partially modified photo-realistic faces using generative models[14].

Among the various attacks, adversarial attacks are the most dangerous because they generally target Deep Neural Networks (DNNs) and focus on Convolutional Neural Networks (CNNs), which are based on the latest FV models. Figure 1 demonstrates the attractiveness of this type of attack and the explosive growth in the number of papers published each year in generating adversarial examples. Therefore, DCNNs are fragile and can be easily attacked by adversarial examples resulting from adding small perturbations[15–17]. These amounts of perturbations are imperceptible to the human eye. The goal is to mislead the classifier to provide wrong prediction outputs because the synthesized images look almost the same as the original ones. Adversarial attacks on FV systems are generated in a manner that humans cannot notice the adversarial perturbations, but the perturbations cause the FV system to misclassify an image, as shown in Fig. 2. Therefore , it is essential to gain a deeper understanding of how these models are susceptible to these attacks.

[1]Computer Science Department, Faculty of Science, Minia University, Al Minya, Egypt. [2]Faculty of Computer Studies, Arab Open University, Muscat, Oman. ✉email: sohair_kilany@mu.edu.eg; ahmed.m@aou.edu.om

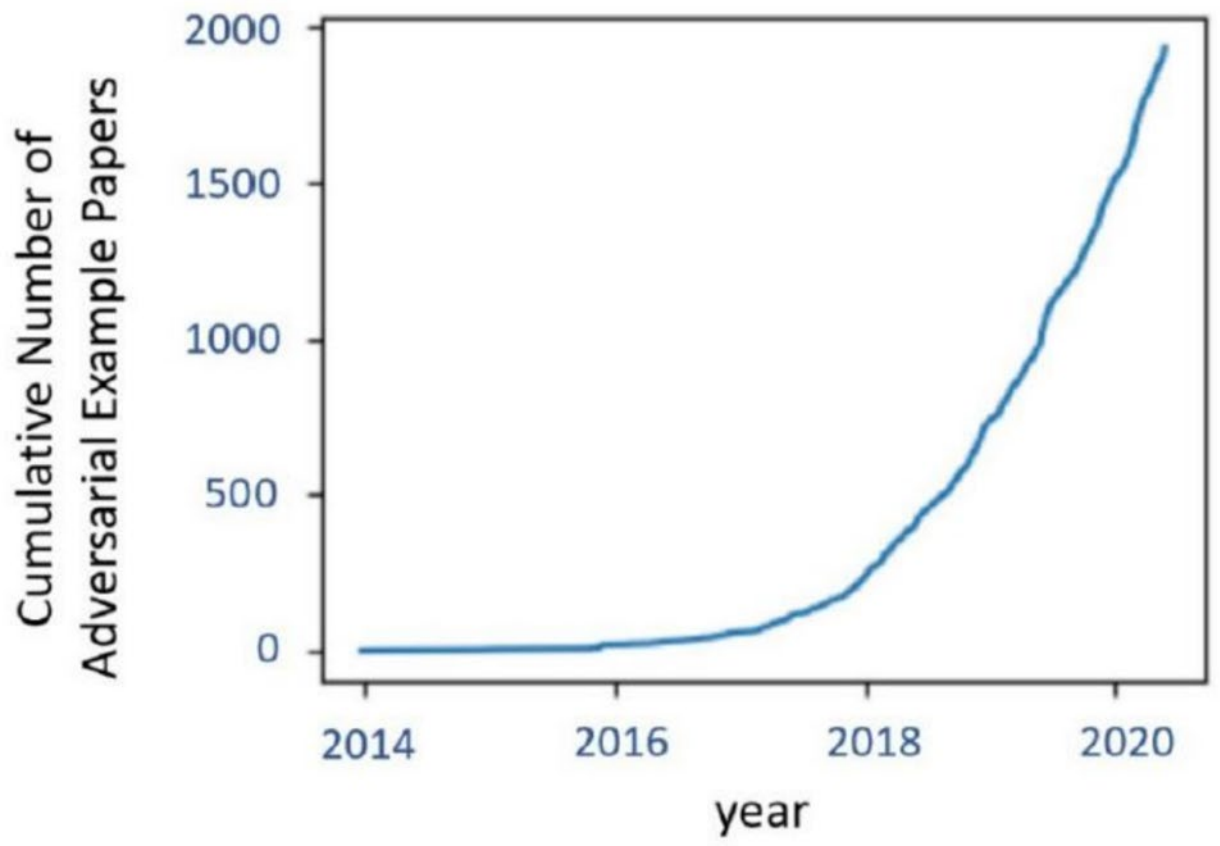

**Fig. 1**. The cumulative number of adversarial example papers published in the last years (Image Credit:[18]).

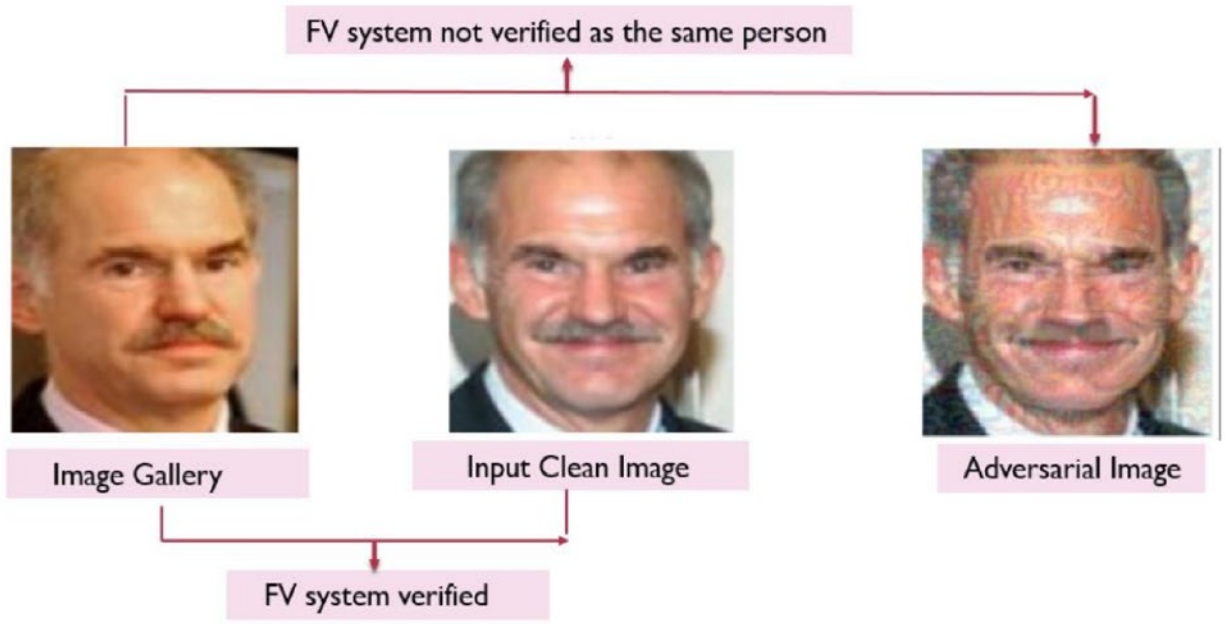

**Fig. 2**. The face verification (FV)system misclassifies the two images belonging to the same person.

Several known methods for crafting adversarial examples vary significantly in terms of their challenges, complexity, computational cost, and the objectives of the attacks. The level of information available to the attacker is classified into three categories (i) White-box attack[19], (ii) Black-box attack[20], and (iii) Semi-white box attack[21]. Depending on the attacker's objective, creating adverse face images can be seen as a security threat. In addition to the apparent security risk of identity theft, this poses an ethical justification to conduct face testing[22]. Another reason to study adversarial attacks on face verification is from the perspective of the machine learning researcher[18]. Identifying new modes of attacks has shown that training on the perturbed instances can help negate the very same line of attacks[23]. Thus, to build robust models, it is necessary to break them up and then adversarially retrain the previously weak model to make it more robust.

Due to the users' privacy issues related to spoofed systems, the requirement to prevent face attacks is becoming increasingly critical. Due to the extensive use of automated face verification systems for border control, failing to detect face attacks might pose a significant security risk. With the introduction of smartphones, we now all have automated facial recognition algorithms integrated into our pockets. Face recognition on our phones makes it easier to (i) unlock the device, (ii) execute financial transactions, and (iii) access privileged content stored on the device.

As a result, considerable literature on defending deep neural networks against adversarial cases has emerged. As indicated in Fig. 3, there are two types of defenses against adversarial attacks. First, Robust Optimization (i.e., Adversarial Training) is the most popular defense method, which modifies the net training procedures or architectures and aims to improve the classifier's robustness[17,23–26]. Although these algorithms are safe against specific attacks, these defenses are still vulnerable to attacks from other mechanisms. However, because online adversarial example generation requires additional computation, adversarial training takes longer than training on clean images alone. Second, the pre-processing strategy leaves the training procedure and architecture unchanged but modifies the data by aiming to detect, remove, or purify the images. For example, in the case of adversarial examples detection, which involves training a binary classifier to distinguish between real and adversarial ones[27–33]. In the case of removing adversarial noise[34,35], which aims to remove the adversarial perturbation by applying transformations as a preprocessing on the input data and then sending these inputs to the target models. But in the case of the purification, it is removed from the input adversarial images only[36] to avoid purifying the real images and consequently avoid the high false reject rates.

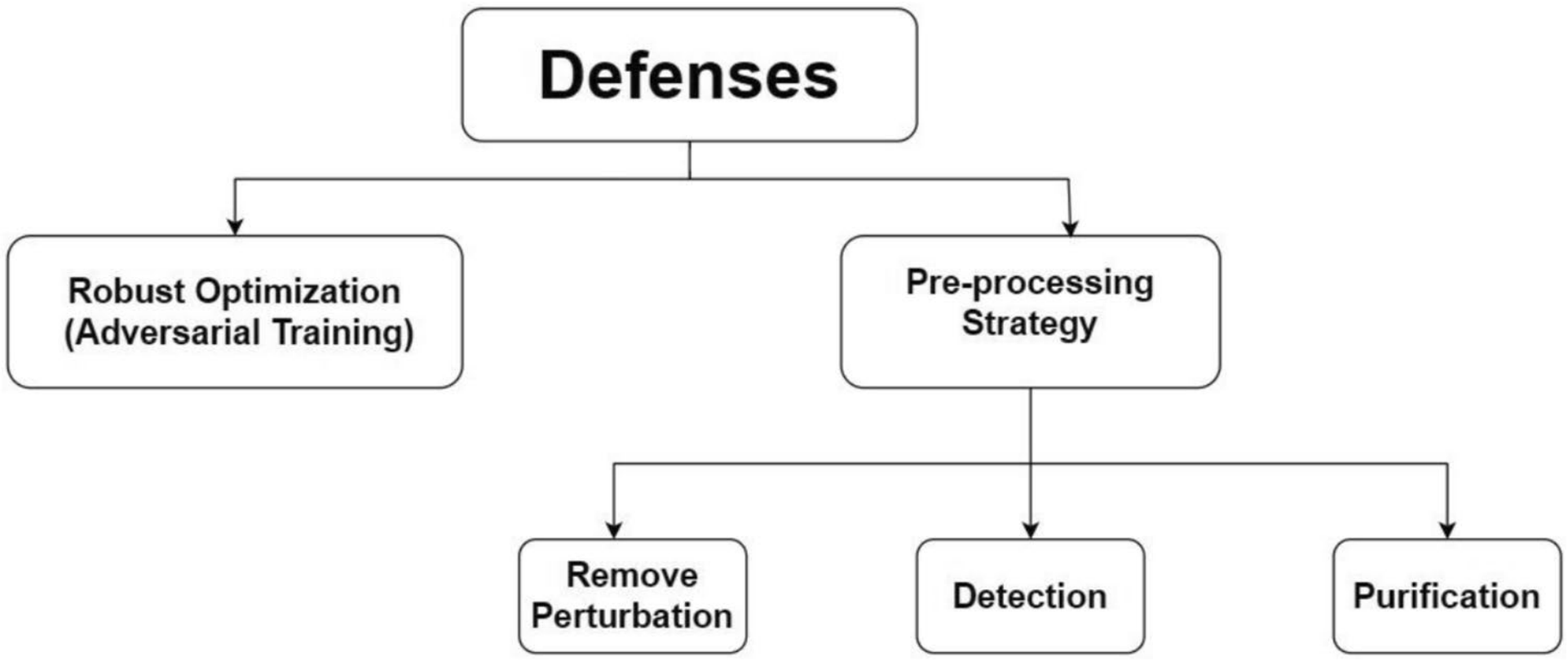

**Fig. 3**. Defense strategies in literature.

The main contributions of this survey are as follows:

- Highlighting the major shortcomings and key areas in facial verification, we present a comparison and analysis of publicly available databases that are vital for both model training and testing.
- Analyzing the state-of-the-art adversarial attacks on FV systems to give an overview of the main techniques and contributions to adversarial attacks.
- Analyzing major defense methods commonly used in the literature and summarizing the most recent related work concerned with detection challenges of Adversarial face images.

The subsequent sections of this paper are structured as follows: Section "Terms and Definitions" introduces pivotal definitions and concepts commonly applied in the realm of adversarial attacks and their corresponding defenses within the framework of the FV system. Section "Face Verification" provides an in-depth examination of the Face Verification system, including diverse network architectures, loss functions, and a comprehensive overview of facial processing algorithms and datasets. Sections "FV Systems Vulnerabilities and Defense" are dedicated to scrutinizing adversarial attack generation methodologies designed to subvert the FV mission and exploring various defensive strategies. The paper concludes in Section "Conclusion".

## Terms and definitions

In this section, we provide a concise overview of the fundamental elements concerning model attacks and defenses within the context of the FV system. Precise terminology definitions play a pivotal role in facilitating comprehension of the primary aspects explored in prior research about adversarial attacks and their corresponding countermeasures. The subsequent sections of this scholarly work consistently adhere to the established definitions of these terms.

### General terms

1. **Adversarial example/image:** A deliberately altered variant of the original image, achieved through the introduction of perturbations such as noise, with the objective of misleading deep convolutional neural networks (DCNN) and machine learning models (ML), including (FV) models.
2. **Adversarial perturbation:** A form of interference is introduced into the clean image to transform it into an adversarial example.
3. **Adversarial training:** A model training procedure incorporating adversarial images in conjunction with unaltered ones.
4. **Transferability:** The capacity of a perturbed example to influence models other than those employed in its creation.

### Specific terms

We can talk about attacks from several aspects, namely: (i) the objective of the attacks and (ii) the information available to the attacker through which the attacks are classified into three categories: (i) White-box attack, (ii) Black-box attack, and (iii) Semi-white box attack.

*Objective of the attacks*

*Poisoning Attack vs Evasion Attack.* Evasion[37] is the most common attack performed during production. It refers to designing an input, which seems normal for a human but is wrongly classified by deep learning models.

A poisoning attack[38] happens when the adversary can inject poisoned data into your model's training, hence getting it to learn something it shouldn't. This attack frequently appears when the adversary has access to the training database.

*Target Attack vs Non-Target Attack:* Indeed, targeted attacks are more complicated than non-targeted ones. The purpose of the non-target attack is to make the model misclassify the adversarial image. In contrast, the targeted attack makes the model classify the adversarial image as a specific target class, which is different from the true class.

*Obfuscation Attack vs Impersonation Attack:* Obfuscation Attacks (OA) and Impersonation Attacks (IA) are frauds. In OA, the attacker seeks to avoid being verified by the FV system, but in IA, the attacker seeks to be incorrectly verified as a different legitimate user by the FV system. In both cases, the fraud process is done by adding an imperceptible perturbation to probe the image. This amount of perturbation is imperceptible to the human eye and differs from one person to another.

*Attacker's information*

- White-Box Attack : In a white-box setting, the attacker has full information about the deep learning models, such as parameters, architecture, defense methods, and gradients . Then, it uses this information to add a small imperceptible perturbation to an inquiry image[17,23]. Perhaps this is not available in the real world for two reasons. First, the attacker cannot access the model because model designers usually do not open their model parameters for special reasons. Second, the model may have been trained in this type of attack, so the model will have the ability to detect this type of attack. As a result of the popularity of the white-box type of attack, designers are creating a robustness model that cannot be deceived by these attacks.
- Black-Box Attack : In a Black-box setting, the attacker does not know the details of target models such as architectures, parameters, and defense methods. They use different models to generate adversarial images, hoping these will transfer to the target model. Additionally, the adversary may have only partial knowledge about (i) the classifier's data domain, for example, handwritten digits, photographs, and human faces, and (ii) the architecture of the classifier, such as CNNs and RNNs.
- Semi-white Box Attack : In a semi-white box attack setting, the attacker trains a generative model for producing adversarial examples in a white-box setting. Once the generative model is trained, the attacker does not need the victim model anymore and can craft adversarial examples in a black-box setting.

## Face verification

Biometrics encompasses technologies designed to authenticate or identify individuals based on physiological attributes or behavioral characteristics. Physiological attributes, such as the iris, fingerprints, and facial features, are generally considered more reliable than behavioral traits like voice patterns, typing style, or walking gait. The domain of deep learning, essential for developing robust verification algorithms, necessitates substantial training data. Obtaining facial images from online sources is notably more straightforward than collecting iris or fingerprint data, primarily due to the widespread availability of digital cameras in affordable smartphones. In contrast, specialized hardware sensors are required for fingerprint and iris recognition. Consequently, the advancement of deep learning techniques has significantly improved face verification performance.

FV technologies are presently deployed in numerous critical commercial and governmental applications. FV plays a pivotal role in preventing identity card duplication, thwarting individuals from obtaining multiple identification documents, such as driver's licenses and passports, under different aliases. Despite the accomplishments of FV systems in the scenarios above, their performance remains constrained, particularly under unconstrained conditions.

As depicted in Fig. 4, factors like varying poses, lighting conditions, facial expressions, age, and obstructions can significantly distort the appearance of an individual's face, highlighting the need to reduce intra-personal variations while accentuating inter-personal distinctions as a central focus in face verification. Moreover, face attacks have materialized in physical realms, such as 3D face masks, and digital domains, encompassing adversarial and digitally manipulated facial images. Malicious actors, called attackers, are increasingly challenging the security of FV pipelines utilized for government services, access control, and financial transactions, often bypassing human operator verification of the legitimacy of facial image acquisition. This section provides an essential foundation for comprehending deep face verification and offers an overview of relevant research in this domain.

In Face Recognition (FR), two primary categories exist: Face Identification (FI), denoted as one-to-many face recognition, and face verification (FV), referred to as one-to-one face recognition. FI is concerned with classifying a face image into a specific identity. At the same time, FV is tasked with ascertaining whether or not two given face images correspond to the same identity. It is important to note that a more efficient FV system directly contributes to the overall efficiency of the FR system. Consequently, face verification can be conceptualized as evaluating the degree of similarity between a pair of facial images.

### Face verification pipeline

Face verification can be simplified as the problem of comparing the similarity between a pair of face images. The whole deep FV system pipeline consists of (i) Input Image (Images or Image Frames), (ii) Face Detector, which localizes faces in images, (iii) Alignment, aligned to normalized canonical coordinates, (iv) Face processing to handle intra-personal variations, (v) Feature Extraction, and (vi) Face Matching. The verification pipeline is shown in Figure 5. The following subsections provide a detailed description of each component of the pipeline.

*Face detector*

One of the most formidable challenges within the field of computer vision pertains to face detection, primarily attributable to the substantial intra-class variability inherent in facial appearances. This variability encompasses factors like skin complexion, background interference, facial orientation, and lighting conditions. Face detection (FD) assumes critical importance within Facial Verification (FV) systems, as it involves precisely localizing and isolating the facial region within an image. FD, in essence, encompasses the capability to identify and delineate one or more faces within a photograph, irrespective of their spatial orientation, lighting variations, attire, accessories, hair color, presence of facial hair, application of makeup, and age of the individuals. In this context, the localization phase involves the placement of a bounding box around the facial region's spatial coordinates within the image. In contrast, the location phase pertains to the precise determination of these coordinates. It is worth noting that classical feature-based methods, exemplified by the "Haar Cascade classifier," have historically provided a reasonably effective solution to this challenge, and such approaches remain popular for face detection[39]. However, recent years have witnessed significant advancements in face detection techniques, with deep learning methodologies emerging as the forefront contenders for achieving state-of-the-art results on established benchmark datasets. Noteworthy examples in this domain include the Multi-task Cascaded Convolutional Neural Network (MTCNN)[40] and RetinaFace[41], which have garnered recognition as leading approaches in the realm of face detection.

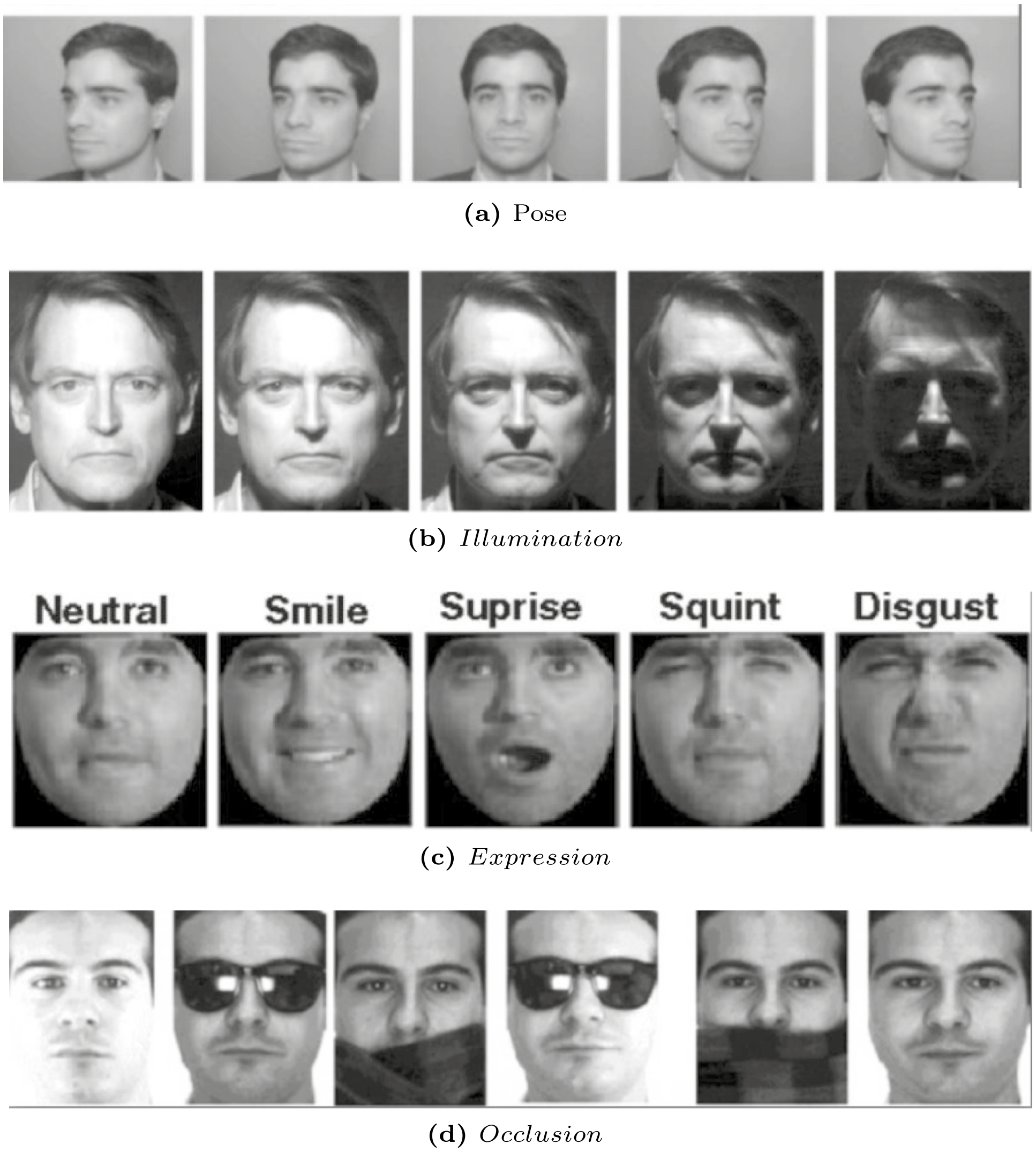

**Fig. 4**. Sources of intra-personal variations: (**a**) pose, (**b**) illumination, (**c**) expression, and (**d**) Occlusion. Each row shows intra-personal variations for the same individual (Image Credit: Google Images).

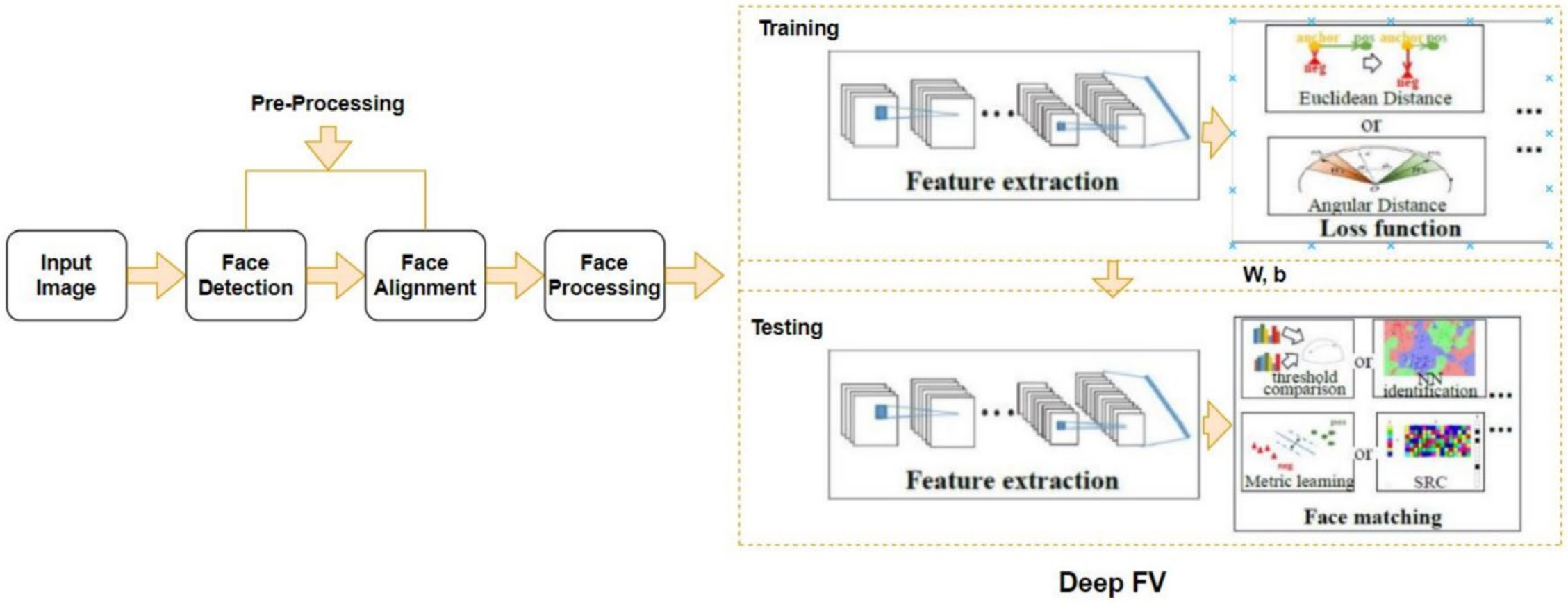

**Fig. 5**. The end-to-end pipeline of the deep FV system.

*Alignment*
Face alignment entails identifying correspondences among facial features, relying on landmark fiducial points, including the eyes, nose, mouth, and jaw. This phase assumes critical importance in the context of face verification. As an illustrative example, Schroff et al.[9] underscored the significance of face alignment following face detection, demonstrating an enhancement in the FaceNet model's accuracy from 98.87% to 99.63%. The simplest approach to alignment involves the application of a basic 2D rigid affine transformation to align the eyes, accounting for variations in facial size and head rotation[9,42]. More advanced techniques employ 3D modeling methods to achieve frontal face alignment. Nevertheless, it is worth noting that 3D face alignment methods are often associated with heightened computational complexity and cost implications.

*Face processing*
The preprocessing stage is frequently regarded as a pivotal phase in constructing machine learning models. Preceding the training and testing phases, it addresses inherent intra-personal variations, encompassing poses, illuminations, expressions, and occlusions. Research conducted by Ghazi et al.[43] conclusively demonstrates that these diverse conditions continue to exert a discernible impact on the efficacy of deep face verification systems. Face-processing methodologies are classified into two distinct categories, namely, (i) one-to-many and (ii) many-to-one approaches.

- One-to-many approaches: These encompass methodologies such as data augmentation, 3D modeling, auto-encoder modeling, and GAN modeling[44–46]. Their primary function is to generate multiple patches or images that encapsulate pose variations derived from a single image. This facilitates the training of deep neural networks in acquiring pose-invariant representations. These strategies address the challenges associated with data acquisition by augmenting training data and expanding the gallery of test data.
- Many-to-one approaches: These include techniques like autoencoder modeling, CNN modeling, and GAN modeling[47–49]. They are designed to transform facial images, specifically by producing frontal views and reducing the variability in appearance within the test data. The objective is to standardize facial alignment and enhance comparability, simplifying face-matching.

*Feature extraction*
In developing an FV system, a critical phase involves extracting a numerical value set referred to as a feature vector or representation. It is imperative to meticulously design the feature vector to prevent the inclusion of superfluous and potentially redundant features, as this can adversely affect verification rates. In recent years, FV systems have been categorized into three distinct approaches for facial representation: (i) holistic, (ii) local, and (iii) shallow and deep learning methods.

- The Holistic Face Representations methodology entails utilizing all pixels present in the input facial image to construct a low-dimensional representation, guided by specific distribution assumptions like the linear subspace[50,51] and sparse representation[52,53]. Nonetheless, it is widely recognized that these theoretically sound, holistic approaches demonstrate limited generalizability when applied to datasets that were not part of their training regimen.
- To build local face representations, face features can also be retrieved from overlapping patches in the face image at several sizes. Local features can be concatenated into a final feature vector summarizing the input face image to add holistic information. The final face representation is usually over-complete, with redundant data and excessive dimensionality. Feature selection, boosting, and dimensionality reduction techniques such as PCA and LDA are utilized to build a more compact face representation. Ahonen et al.[54] proposed Local Binary Patterns (LBP) for face recognition. They divide the face image into a grid to exploit local and global facial features. A histogram of LBP characteristics is generated for each cell in the grid, and the resulting face representation is concatenated.
- In facial verification, Convolutional Neural Networks (CNNs) have outperformed human capabilities on various benchmark assessments[10]. The proliferation of extensive facial datasets and improved computational resources, notably Graphics Processing Units (GPUs), has led to a notable surge in interest over the past few years in automatic feature extraction techniques centered around Convolutional Neural Networks (CNNs)[10], as evidenced by the data in Table 1. Diverse architectural configurations and loss functions have been employed to extract distinguishing identity attributes from facial images through the utilization of Deep Convolutional Neural Networks (DCNNs). This has been achieved by carefully designing loss functions to augment discriminative capacity during the training process[55].

| Method | Loss | Architecture | Training Set | LFW | YTF |
|---|---|---|---|---|---|
| DeepFace[10] | Softmax | AlexNet | Facebook SFC (4.4M, 4K) | 97.35% | 91.4% |
| FaceNet[9] | Triplet | Inception | Google (200M, 8M) | 99.63% | 95.12% |
| VGGFace[57] | Softmax | VGG-16 | VGGFace (2.6M, 2.6K) | 98.95% | 97.3% |
| SphereFace[7] | A-Softmax | ResNet-64 | CASIA-WebFace (0.49M, 10K) | 99.42% | 95.0% |
| ArcFace[8] | ArcFace | ResNet-100 | MS1M (5.8M, 85K) | 99.83% | 98.02% |
| Gate-FV[58] | Angular | MDCNN | CASIA-WebFace (0.49M, 10K) | 99.38% | 94.3% |

**Table 1**. Face verification methods evaluated on LFW and YTF datasets.

| Database Name | # Images | # Identities | Availability | Type |
|---|---|---|---|---|
| LFW[11] | 13,233 | 5,749 | Public | Test |
| YTF[12] | 3,425 videos | 1,595 | Public | Test |
| Facebook[10] | 4.4M | 4K | Private | Train |
| CASIA-WebFace[59] | 494,414 | 10,575 | Public | Train |
| Google-FaceNet[9] | 200M | 8M | Private | Train |
| VGGFace[57] | 2.6M | 2.6K | Public | Train |
| IJB-A[60] | 5,712 | 500 | Public | Test |
| MegaFace[61] | 4.7M | 672,057 | Public | Train |
| VGGFace2[62] | 3.31M | 9,131 | Public | Train |
| CPLFW[63] | 11,652 | 3,968 | Public | Test |
| WebFace260M[64] | 260M | 4M | Public | Train |

**Table 2**. Common face verification datasets

One of the primary challenges faced by the facial verification (FV) research community was the one-shot learning problem. This challenge involved the development of FV systems capable of verifying a person's identity using just a single example of their face. Historically, deep learning algorithms struggled with this scenario, as discussed in reference[56]—however, DeepFace[10] successfully addressed this issue and achieved an impressive accuracy rate of 97.35% on the Labeled Faces in the Wild (LFW) benchmark dataset, approaching human-level performance. Additionally, it significantly reduced the error rate of the YouTube Face database by more than 50%. This achievement was accomplished by training a nine-layer deep neural network (CNN) with over 120 million parameters on a dataset containing four million facial images with 3D alignment for face processing.

Furthermore, an end-to-end metric learning approach was tested using a Siamese neural network replicated twice during training. This network takes two images as input and outputs the degree of difference between their features, followed by a top fully connected layer that maps this information into a single logistic unit. Schroff et al.[9] introduced FaceNet, which utilized a triplet loss function to learn a Euclidean distance metric for measuring face similarity. This approach achieved remarkable accuracies of 99.63%±0.09 with additional face alignment, 98.87%±0.15 when using a fixed center crop on LFW, and 95.12% on the YouTube Faces dataset.

Notably, these methods emphasize end-to-end learning, representing the entire system directly from facial pixels rather than relying on engineered features. They also require minimal alignment, typically focusing on a tight crop around the facial region[9].

*Face matching*

In the context of facial feature extraction, the primary objective of an FV system is to determine the degree of similarity between these extracted features. This is achieved by applying similarity measurement techniques, with commonly used methods including cosine similarity[8] and Euclidean distance[9]. While Euclidean distance is a straightforward choice for comparing feature vectors, other distance metrics such as cosine similarity, Manhattan distance, histogram intersection, log-likelihood statistics, and chi-square statistics have been explored to enhance face verification performance.

## Benchmark dataset

In recent years, a discernible pattern has arisen, characterized by a transition from small-scale to large-scale experimentation, a shift from reliance on single sources to incorporating diverse sources, and a progression from laboratory-controlled settings to unconstrained real-world conditions. Table 2 presents a comprehensive compilation of data about a collection of benchmark datasets utilized within the academic literature. This compilation encompasses various elements, including database size quantified by the number of images, the presence of identifiable faces, and the intended applications.

- Faces in the Wild (LFW)[11] stands as one of the pioneering databases specifically tailored for the investigation of uncontrolled, “in-the-wild” face verification scenarios. This repository encompasses a considerably larger volume of images, which serve as essential evaluative material for algorithms designed for practical, real-world applications. Within the domain of face verification, LFW continues to hold its status as a key benchmark. The inception of this dataset dates back to its initial release in 2007[11], and subsequently, it underwent updates in 2014[65]. It comprises 13,233 facial images, each sized at 250x250 pixels, representing 5,749 distinct individuals, with 4,069 of these individuals featuring in only a single image.
- YouTube Faces (YTF)[12] The YTF dataset was curated to investigate unconstrained videos featuring matched background similarities. Comprising 3,425 videos drawn from 1,595 distinct subjects, the dataset exhibits an average of 2.15 videos per subject. Notably, video durations within the dataset range from 48 frames for the shortest clip to 6,070 frames for the longest, with an average video clip length of 181.3 frames.
- IARPA Janus Benchmark A (IJB-A)[60] encompasses a heterogeneous collection of visual data, including both images and videos, originating from a pool of 500 subjects captured in diverse real-world scenarios. Notably, for each subject included in the dataset, a minimum of five images and one video is available. The IJB-A dataset comprises 5,712 images and 2,085 videos, translating to an average of 11.4 images and 4.2 videos per subject. At the granularity of individual subjects, the dataset is structured into ten distinct, randomly generated training and testing splits, each encompassing all 500 subjects in IJB-A. For each of these splits, a subset of 333 subjects is randomly allocated to the training set, serving as a foundation for algorithmic model development and acquiring insights into facial variations germane to the Janus challenge. The remaining 167 individuals are assigned to the testing set for evaluation and validation.
- Cross-Pose LFW (CPLFW)[63] database represents a revitalized iteration of the Labeled Faces in the Wild (LFW) dataset, which serves as the prevailing benchmark for evaluating unconstrained face verification algorithms. Within the LFW framework, ten distinct sets of image pairs have been meticulously constructed for cross-validation, each containing 300 positive and negative pairs. These subsets are organized based on unique subject identities, ensuring each identity is exclusively featured in a single subgroup. In contrast, the CPLFW dataset employs an analogous partitioning approach, creating ten partitions or “folds,” mirroring the identity distributions found in the original LFW folds. Notably, each individual within the CPLFW dataset is represented by a set of two to three images.
- MegaFace[61] stands as a substantial publicly available face recognition training dataset that has established itself as an industry benchmark. Within MegaFace, a comprehensive compilation of 4,753,320 facial images is available , representing a diverse pool of 672,057 distinct identities sourced from a repository of 3,311,471 photographs from the personal albums of 48,383 users on the Flickr platform. Notably, while the photos featured in MegaFace predominantly possessed Creative Commons licenses, most did not permit commercial usage.
- CASIA-Webface[59] dataset is a valuable resource for addressing face verification and recognition challenges. Comprising a collection of 500,000 facial images featuring 10,575 distinct celebrities, these images were sourced from publicly available online sources, capturing subjects in uncontrolled, real-world settings, thus characterizing the “in the wild” nature of the dataset acquisition process.
- VGGFace2[62] consists of 3.3 million facial images of celebrities drawn from a pool of 9,000 unique identities, with an average of 362 images available per subject, the creators of the dataset prioritized the meticulous reduction of label inaccuracies, alongside the deliberate inclusion of a wide spectrum of facial poses and age groups. These meticulous efforts have rendered the VGGFace2 dataset an optimal selection for training advanced deep-learning models designed to excel in tasks related to facial analysis.
- VGGface[57] dataset comprises an extensive compilation of 2.6 million facial images, encompassing 2,622 unique identities. Each identity is accompanied by an associated text file with image URLs and corresponding facial detection information.

## Evaluation metrics

The facial matching procedure involves the computation of the dissimilarity measure between a given pair of facial images, which is subsequently compared to a predefined threshold. When the calculated dissimilarity measure falls below this threshold, the pair of faces is classified as belonging to the same individual; otherwise, they are deemed to represent distinct individuals. This categorization identifies correctly matched pairs as either true positives (indicating same-person pairs) or true negatives (indicating different-person pairs). Within this context, two types of errors may occur: (i) false positives, also known as false acceptances, correspond to instances where different individuals are erroneously identified as the same person, and (ii) false negatives, or false rejections, occur when the same individual is mistakenly categorized as distinct individuals. The assessment of facial verification performance relies on the evaluation of these two error types, with the utilization of the subsequent metrics:

Accuracy represents the percentage of truly recognized pairs, both positive and negative.

$$\tau = \frac{Number\ of\ successful\ pairwise\ matches}{Total\ number\ of\ image\ pairs} \tag{1}$$

where the numerator represents the Number of sucessful pairwise matches and the denominator represents the Total number of image pairs. Verification Accuracy ($\tau$) represents the acceptance threshold and is determined using cross-validation.

We also report the True Accept Rate at a pre-determined False Accept Rate. The $\tau$ is determined via a Receiver Operating Characteristic (ROC) curve. Formally,

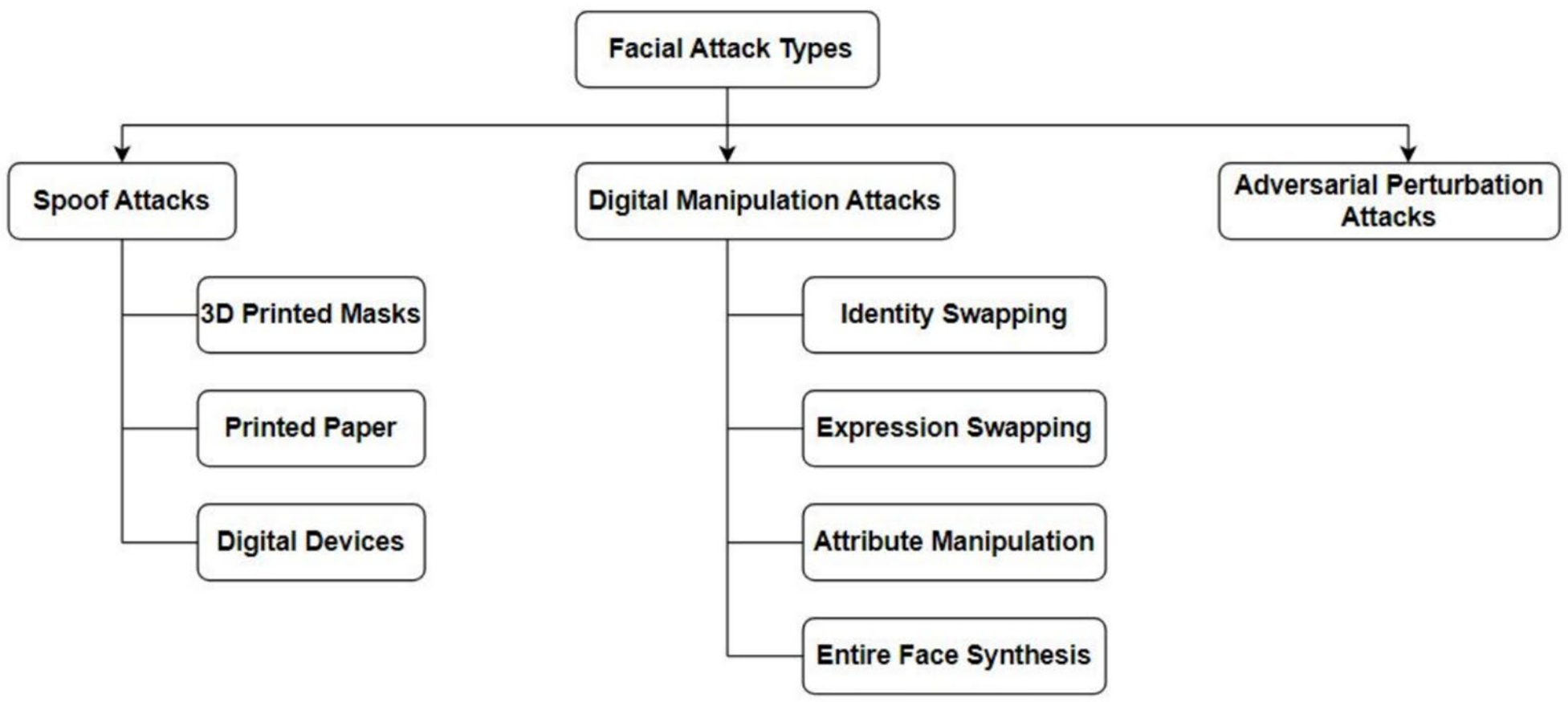

**Fig. 6**. The broad categorization of facial attack types aimed to deceive the FV systems.

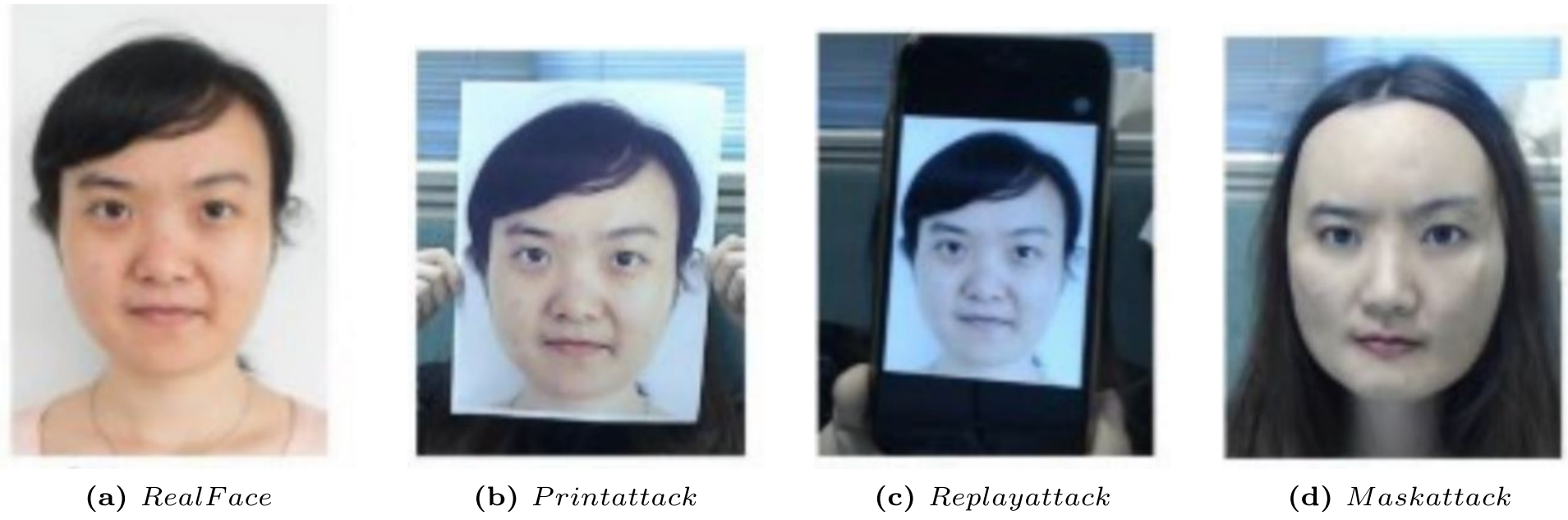

**Fig. 7**. Face presentation attacks require a physical artifact. (**a**) real face, from (**b-d**) represents three types of face presentation attacks: (**b**) printed photograph, (**c**) replaying the targeted person's video on a smartphone, and (**d**) a 3D mask of the target's face (Image Credit: Google Images).

$$TAR\ (\tau) = \frac{Number\ of\ genuine\ pairs\ with\ similarity\ score\ >\ \tau}{Total\ number\ of\ genuine\ pairs} \tag{2}$$

$$FAR\ (\tau) = \frac{Number\ of\ impostor\ pairs\ with\ similarity\ score\ >\ \tau}{Total\ number\ of\ impostor\ pairs} \tag{3}$$

Equal Error Rate (EER) is the error when false positive rate (FPR) and false negative rate (FNR) are the same, which is found by varying the threshold. Receiver Operating Characteristic (ROC) is the curve of true positive rate (TPR) against false positive rate (FPR) that is calculated by varying the threshold.

## FV systems vulnerabilities

Despite the impressive verification performance achieved using deep learning models, the FV systems remain vulnerable to the growing threat of face attacks, such as face spoofing and adversarial perturbations, in both the physical and digital domains. For example, an attacker can hide his identity by using a printed photograph, a worn mask[66], or even an image displayed on another electronic device to present a fake face to the biometric sensor, or intruders can assume a victim's identity by digitally swapping their face with the victim's face image[14].

There are three types of facial attacks depicted in Fig. 6: (i) Spoofing attacks: physical domain artifacts such as 3D masks, eyeglasses, and replaying videos[13], (ii) Adversarial perturbation attacks: imperceptible noises added to probes to evade FV systems, and (iii) Digital manipulation attacks: entirely or partially modified photo-realistic faces using generative models[14]. There are various attack types in each of these categories. For example, 13 common types of spoofing attacks[13]. Similarly, in adversarial and digital manipulation attacks, each attack model is designed with unique objectives and losses and can be regarded as one attack type.

Several face-attack defense approaches have been proposed to protect FV systems from these attacks. Because the exact type of face attack may not be known a priori, a generalizable defense that can defend an FV system against any of the three attack categories is critical.

- Spoof Attacks: are presentation attacks targeting facial recognition systems, as illustrated in Fig. 7, encompass various physical counterfeiting techniques necessitating active engagement by actors. These methods encompass the utilization of tangible counterfeits like 3D printed masks, printed images on paper, or digital tools such as video replays on mobile devices, all of which enable the impersonation of an individual's identity or the concealing of the attacker's identity. The ease with which an assailant can employ these tactics, such as displaying videos featuring the victim's visage or submitting printed representations of the victim to a Facial Verification (FV) system[67].

Even if a countermeasure system were in place, leveraging depth sensors to detect face presentation attacks, it would still be susceptible to more advanced subterfuge techniques. Attackers may resort to using 3D masks[68], cosmetic disguises, or even virtual reality simulations[69], thereby enabling the execution of more intricate and sophisticated attacks.

- Adversarial Perturbation Attacks: Most facial verification (FV) models are predominantly constructed using Deep Convolutional Neural Networks (DCNNs) and have consistently demonstrated impressive performance and high accuracy in recent years. Nonetheless, DCNNs exhibit susceptibility to adversarial examples, which are generated through minor perturbations introduced into input samples[15–17]. Adversarial perturbations, exemplified in Fig. 8, can be defined as minimal alterations represented by $\epsilon$, where the addition of this perturbation to the input image x, denoted as $(x + \epsilon)$, results in the misclassification of the input by deep learning models. Despite the imperceptibility of the perturbation to the human eye, it constitutes an adversarial example in image classification, capable of causing CNNs to misclassify the image. According to certain seminal works, such as[70], the emergence of adversarial examples can be attributed to the limited generalization capabilities of DNN models, possibly stemming from the high complexity of their architectural design. The investigation into the existence of adversarial examples holds significance, as it can provide valuable insights for designing more robust models and enhancing our comprehension of existing deep learning frameworks.

  Malicious actors can manipulate their facial images to deceive FV systems, leading to two primary types of attacks: impersonation attacks, where the attacker aims to be recognized as a target victim, and obfuscation attacks, where the attacker seeks to be matched with a different identity within the system. However, the adversarial facial image generated by such attacks should appear legitimate to human observers. In contrast, face presentation attacks, as depicted in Fig. 7, involve the attacker physically presenting a fake face as the target identity to the FV system. These attacks are typically more conspicuous to human observers, especially in situations involving human operators, such as those in airports.

  Notably, adversarial examples possess the characteristic of transferability, meaning that adversarial examples created to target one specific victim model are also highly likely to mislead other models. This property of transferability is often exploited in black-box attack techniques[71]. If attackers obscure the model's parameters, they can resort to attacking alternative models, thereby leveraging the portability of the generated adversarial samples. Defense methods also harness the property of portability, as demonstrated in[72], by utilizing adversarial training with samples created to perturb one type of model to bolster the defense of another type of model.

  In this section, our primary focus is on exploring adversarial perturbations, as detailed in Section 4.1.
- Digital Manipulation Attacks: is the process of affording the capacity to comprehensively or partially alter genuine facial images using Variational Auto Encoders (VAEs) and Generative Adversarial Networks (GANs)[14]. Figure 9 shows different face presentation attacks based on digital manipulation. These digital manipulation attacks may be categorized into distinct types, as outlined below:

***Identity Swapping:***

These methods replace one person's face with another person's face digitally. For example, FaceSwap[73] contains well-known actors in movie scenes in which they have never been featured before. DeepFakes also uses deep learning algorithms to produce face swaps.

***Expression Swapping:***

These methods exchange expressions in real time using only RGB cameras. Expressions in the facial image can be digitally and artificially replaced by others[74].

***Attribute Manipulation:***

studies like StarGAN[75] and STGAN[76] use the latest GANs to manipulate attributes by changing single or multiple traits in a facial image, such as gender, age, skin color, hair, and glasses.

***Entire Face Synthesis:***

An attacker can easily synthesize entire facial images of unknown identities, which are so realistic that even humans have difficulty assessing whether they are authentic or manipulated[77]. Due to the advent of GANs and large-scale, high-resolution facial data sets.

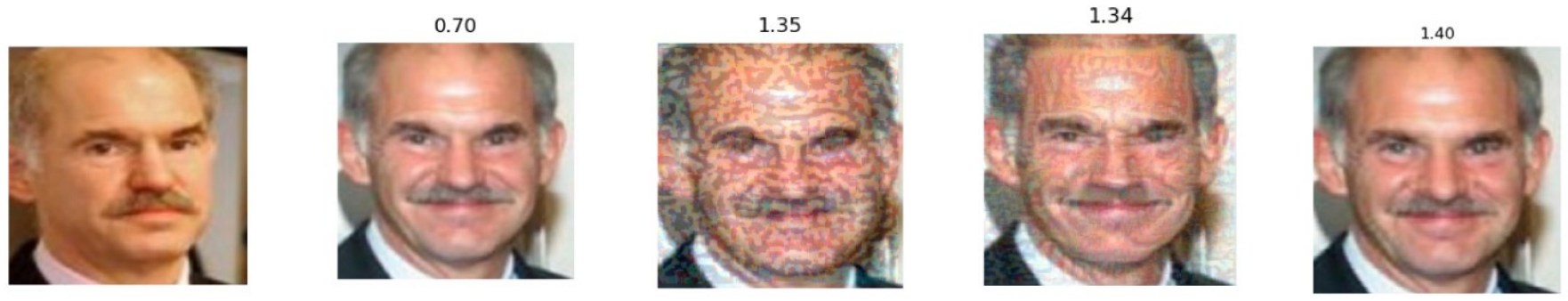

(a) *EnrolledFace* (b) *InputProbe* (c) *FGSM* (d) *PGD* (e) *AdvFaces*

**Fig. 8**. Image samples for probe face images and their corresponding synthesized ones. (**a**) gallery enrolled images and (**b**) probed images for the same person (**c**) FGSM (**d**) PGD (**e**) AdvFaces. Euclidean distance scores were obtained by comparing (**b-e**) to the enrolled images.

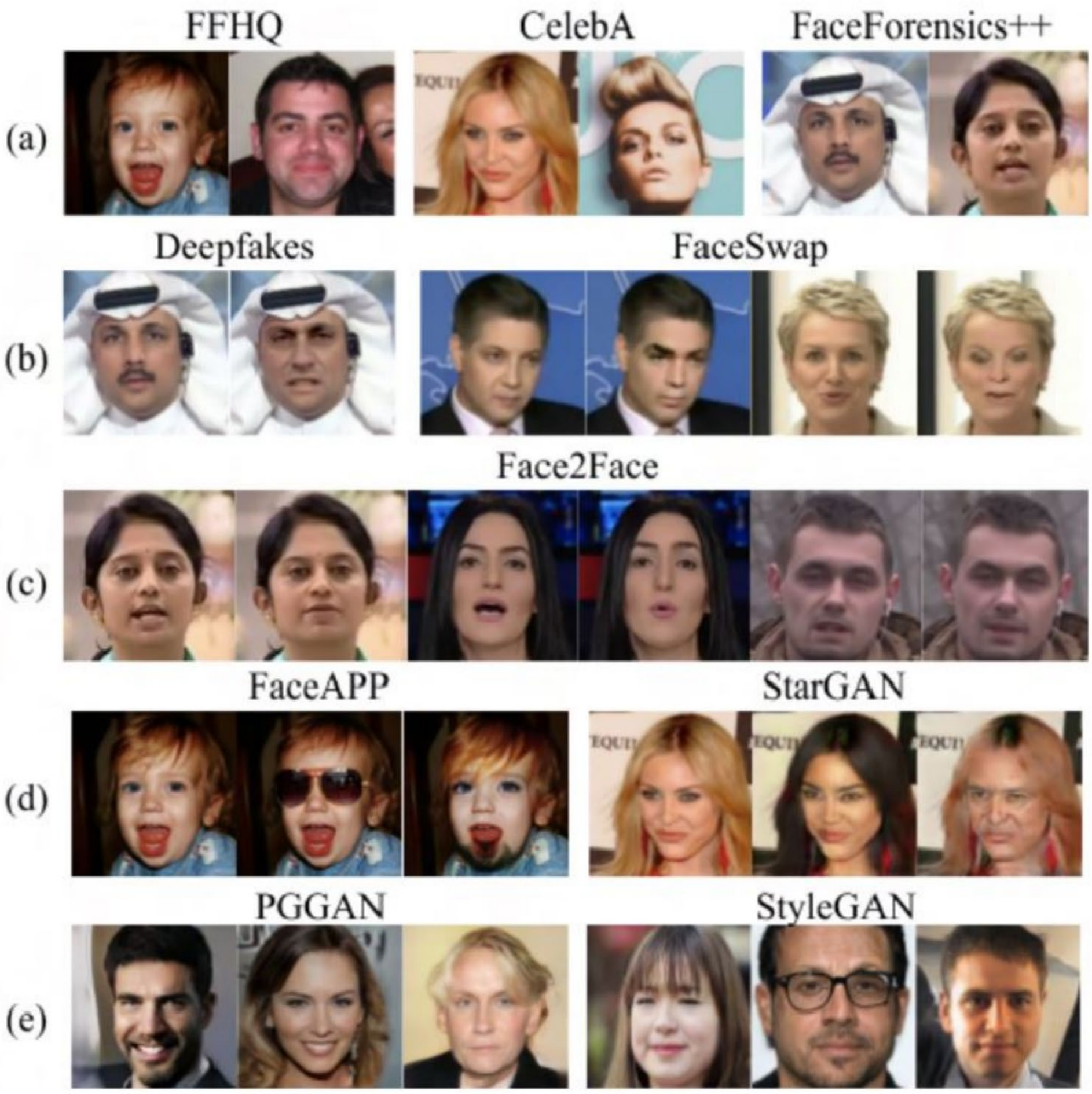

**Fig. 9**. Examples of digitally manipulated faces from different sources such as FFHQ, CelebA, FaceForensics++, FaceAPP and StarGAN, PGGAN, StyleGAN datasets (Image Credit:[14]).

## Adversarial example generation

In the realm of computer vision, a multitude of techniques for generating adversarial examples have been developed. These methodologies aim to introduce subtle perturbations into specific images, thereby inducing erroneous classifications by machine learning models[17,78]. To formalize the concept of adversarial examples, we introduce the following notation: let x represent an input that is accurately classified as y, and f(x) denote the classification decision made by the machine learning model. An adversarial example, denoted as $x^{adv}$ = x + $\epsilon$, is formed through the addition of a perturbation $\epsilon$, satisfying the condition:

$$f(x) = y \ and \ f(x^{adv}) \neq y \tag{4}$$

Generating adversarial examples ultimately involves identifying perturbations in input data that remain imperceptible to human observers yet lead to misclassifications by vulnerable machine learning models. Additionally, researchers have demonstrated the creation of a general perturbation applied to a dataset, resulting in the high likelihood of misclassification for numerous normal images[16]. Szegedy et al.[70] unveiled the existence of adversarial examples and introduced the first algorithm capable of reliably detecting adversarial perturbations. The Fast Gradient Sign Method (FGSM), proposed by Goodfellow et al.[17] and illustrated in Fig. 10, forms the basis of this algorithm. Kurakin et al.[25] extended the FGSM method to generate larger quantities of adversarial examples, enhance the quality of the generated adversarial models, and enable the execution of targeted attacks. However, it is worth noting that some of these extensions come at the expense of increased computational resources.

Moosavi-Dezfooli et al.[16] introduced Deepfool, designed initially for untargeted attacks with a focus on improving perturbations in the $L_2$ norm but adaptable to any $L_p$ norm. Their approach is highly effective and can discover smaller perturbations compared to Szegedy et al.'s L-BFGS approach[70]. The fundamental idea behind their proposed algorithm is an iterative approximation of the hyperplanes that separate distinct classes and the distances between perturbed inputs and decision boundaries, estimated through orthogonal projections. While Deepfool cannot guarantee the discovery of the optimal solution with the minimal perturbation for a given input, the authors assert that the resulting perturbation approximates the minimal perturbation effectively.

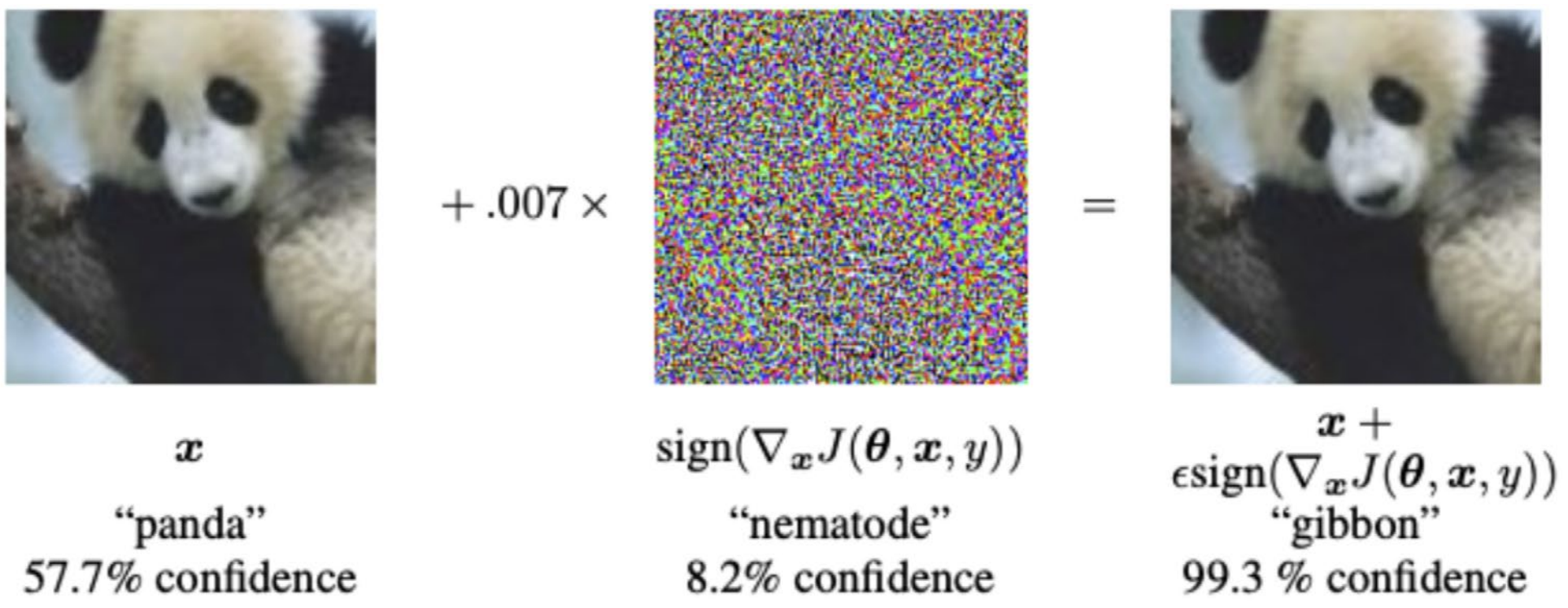

**Fig. 10**. By adding small perturbations (distortions) to the original image, which results in the model labeling this image as a gibbon, with high confidence (Image Credit:[17]).

| Attack Method | Attack Settings | Similarity Metric | Scope | Domain | Attack Objectives |
|---|---|---|---|---|---|
| FGSM[17] | White-Box | $L_\infty, L_2$ | Universal | Classification | Obfuscation |
| Face Recognition[19] | White-Box | Physical | Image-specific | Recognition | Impersonation |
| PGD[23] | White-Box | $L_\infty$ | Universal | Classification | Obfuscation |
| A3GN[84] | White-Box | Cosine | Image-specific | Recognition | Impersonation |
| Evolutionary Optimization[20] | Black-Box | – | Image-specific | Recognition | Impersonation |
| GFLM[85] | White-Box | – | Image-specific | Recognition | Obfuscation |
| AdvFaces[21] | Semi-White-Box | $L_2$ | Image-specific | Recognition | Both |
| GAP++[88] | White-Box | $L_\infty, L_2, L_0$ | Universal | – | Obfuscation |

**Table 3**. Comparison of different adversarial attack methods.

Papernot et al.[79] introduced the Jacobian-based Saliency Map Attack (JSMA), optimized for the L0 distance metric. This method leverages the gradients of the records or softmax units with respect to the input image to compute a saliency map, which approximates the impact of each pixel on the image's classification. JSMA perturbs the most significant pixels until a targeted attack succeeds or the number of perturbed pixels exceeds a predefined threshold.

Carlini and Wagner[80] proposed three algorithmic models for generating optimized adversarial examples with $L_0$, $L_2$ or $L_\infty$ norms. All three variants can perform targeted attacks by minimizing the respective objective functions, which can also govern prediction confidence. Finally, Sabbour et al.[81] focus on directly manipulating deep representations through imperceptibly small perturbations instead of inducing explicit misclassifications. While the previously mentioned methods aim to create adversarial models leading to misclassified labels, this approach seeks to transform the internal representations of an image to closely resemble those of an image from a different category with imperceptible perturbations.

Recent methods have focused on improving the transferability of adversarial examples, ensuring effectiveness even when attackers have no direct knowledge of the target model. Zhou et al.[82] introduced the Diverse Parameters Augmentation (DPA) method, which diversifies surrogate models by training multiple intermediate checkpoints with varied parameter initializations, significantly enhancing adversarial transferability for face recognition tasks. Complementarily, research by Yu et al.[83] has shown that embedding carefully designed triggers within adversarial examples (trigger activation techniques) can effectively evade traditional defensive mechanisms, further increasing the robustness and transferability of generated adversarial examples.

## Adversarial attacks on image classification

In image classification, adversarial examples refer to intentionally crafted images that closely resemble their original counterparts but can elicit erroneous predictions from the classifier. These adversarial perturbations must be imperceptible to human observers. Therefore, the investigation of adversarial examples within the domain of images holds significant importance for two primary reasons: (a) the perceptual similarity between counterfeit and genuine images is readily discernible to human observers, and (b) the structural simplicity of both image data and image classifiers, in contrast to other domains such as graphs or audio, has led to numerous studies treating image classifiers as a standard case.

Many adversarial example generation methods have been proposed in recent years, as documented in Table 3. As evidenced in previous research, many of these attack methods can be categorized into intensity-based attacks[17] and geometry-based attacks[84,85]. For instance, Goodfellow et al.[17] introduced a white-box attack named the Fast Gradient Sign method (FGSM), which introduces subtle perturbations to various regions of the original image via back-propagation through the target model, causing the model to confidently misclassify the adversarial image into another category, as illustrated in Fig. 10. However, their approach is associated with several drawbacks, such as the excessive computational time required for generating adversarial examples and the resultant degradation in the perceptual quality of the generated images. Additionally, it relies on the white-box attack paradigm, which proves impractical in real-world scenarios. Moreover, the requirement of applying perturbations to all regions of the image and relying on softmax probabilities for evading an image classifier is not viable, especially in cases like the FV system, where the classifier does not employ a fixed set of classes (identities). Moosavi-Dezfooli et al.[16] proposed image-agnostic adversarial attacks, which entail the generation of universal perturbations capable of deceiving the classifier across a wide range of image types and models. Recent work also shows that backdoor-style triggers, e.g., universal 'master-key' patterns, can force a network to verify an impostor as the genuine user[86].

In image classification scenarios, adversarial threats have evolved towards more stealthy and practical black-box attack strategies. For instance, Park et al.[87] proposed the *Mind the Gap* technique, which analyzes incremental query updates to systematically craft adversarial images under black-box conditions. This approach demonstrates how iterative query strategies can bypass detection mechanisms, underscoring the growing complexity and sophistication of adversarial attacks on image classification models.

## Adversarial attacks on face verification (FV) systems

Utilizing deep learning models, the Facial Verification (FV) system can attain a noteworthy True Accept Rate (TAR) of 99.27% while maintaining an impressively low False Accept Rate (FAR) of 0.001% with genuine face pairs when leveraging FaceNet[9]. This level of performance is attributed to the ample availability of extensive facial datasets for training these models and the incorporation of Convolutional Neural Network (CNN) architectures, as illustrated in Table 1. However, it should be noted that CNN models are susceptible to adversarial perturbations, as elucidated in Table 3. Even minute imperceptible perturbations, undetectable to the human eye, can lead to misclassification by the CNN[70].

Notwithstanding their commendable verification capabilities, mainstream FV systems are still exposed to an escalating risk posed by adversarial examples, as depicted in Figure 8. To compromise the FV system, an adversary can intentionally manipulate their facial image to deceive the FV system into incorrectly identifying them as the intended target (impersonation attack) or as a different individual (obfuscation attack). Crucially, the manipulated facial image must convincingly appear as a legitimate representation of the adversary to human observers, as illustrated in Fig. 11.

However, it is essential to note that in the context of adversarial faces, the adversary need not actively engage in the authentication process when comparing their probe and gallery images. Conversely, in scenarios involving presentation attacks, such as the use of masks or the replay of images/videos of genuine individuals, the adversary must actively participate. Such active participation may be discernible in situations involving human operators. Hence, it is imperative to thoroughly investigate the spectrum of attacks to which CNN models are susceptible to comprehensively assess their vulnerabilities and limitations. This understanding should inform the design and modification of CNN models by developers in the future.

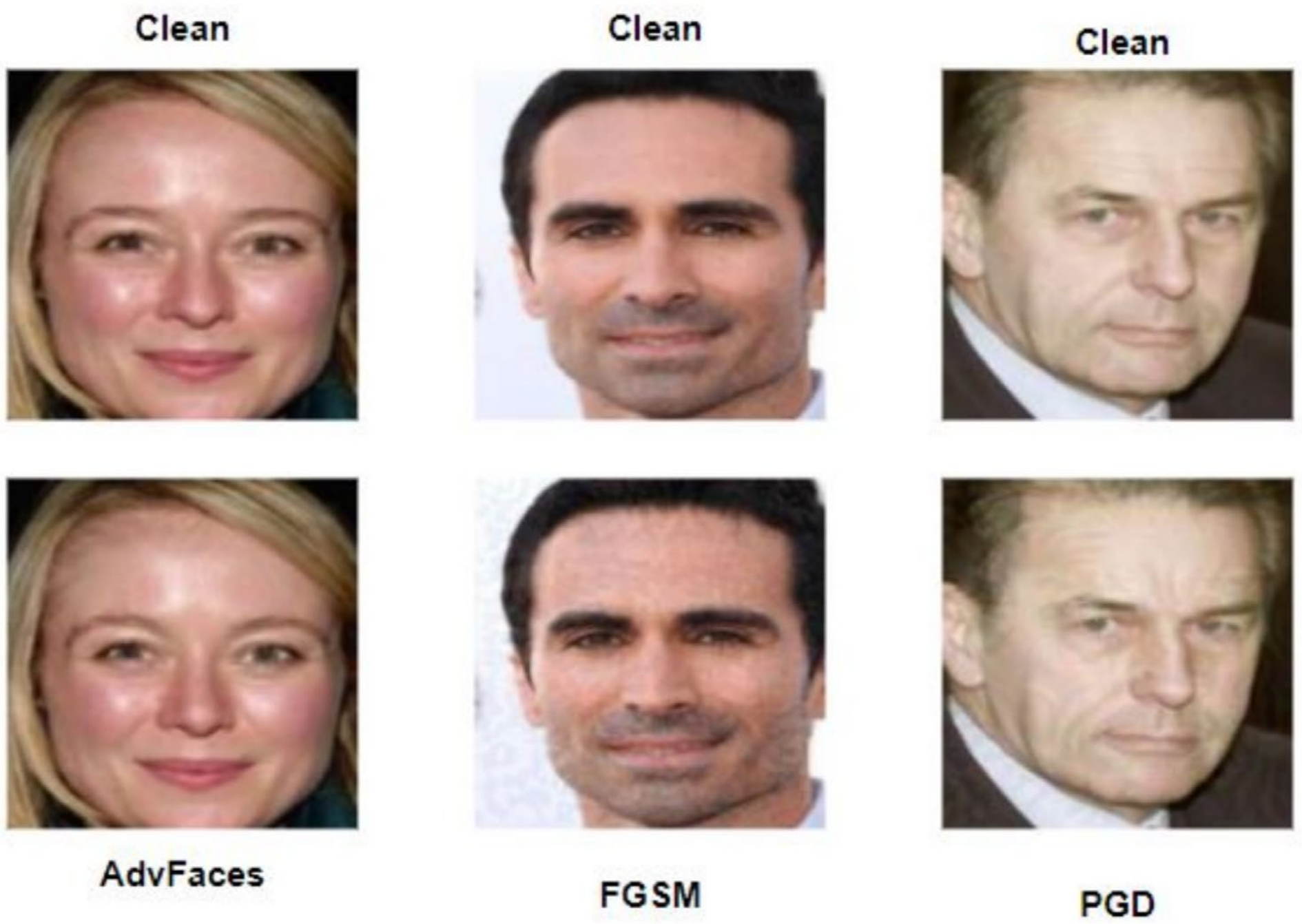

**Fig. 11**. Examples of adversarial Attacks: (**a**) AdvFaces, (**b**) FGSM, (**c**) PGD.

Recently, a prominent area of research has emerged in the field of face verification, focusing on the generation of adversarial examples. Researchers such as Bose et al.[89] have pursued this inquiry by employing constrained optimization techniques to create adversarial examples that elude face detection systems. Similarly, Dong et al.[20] have introduced an evolutionary optimization approach for generating adversarial faces, particularly in black-box scenarios. Nevertheless, this method necessitates a substantial number of queries to yield satisfactory results.

Song et al.[84] have contributed to this discourse by proposing an attention-driven adversarial attack generative network (A3GN) tailored for producing counterfeit face images within a white-box framework, emphasizing impersonation attacks. However, their method exhibits certain limitations, including a requirement for access to gallery-enrolled face images, which may be impractical in real-world settings. Furthermore, it mandates at least five images of the target individual for training and is confined to targeting a single subject. The image generation process is characterized by low-quality and time-consuming operations.

Many of these shortcomings have been addressed by Deb et al.[21], who introduced a semi-white box attack known as 'Advfaces.' This method generates adversarial examples by feeding the network and necessitates only a single face image of the target subject for both training and inference. Importantly, Advfaces produces adversarial images of high quality and perceptual realism , posing no discernible threat to the human eye. Its efficacy is demonstrated by its ability to outperform state-of-the-art face matches, achieving an impressive attack success rate of 97.22% in obfuscation attacks. Moreover, Advfaces exhibits the property of transferability, wherein adversarial examples designed to target one victim model have a high likelihood of confounding other models.

We provide a succinct overview of the three popular obfuscation adversarial attacks FGSM[17], PGD[23], and Advfaces [21], on face verification systems, elucidating the process by which the magnitude of perturbation is precisely determined.

- Fast Gradient Sign Method (FGSM)[17]: FGSM is a fast method to generate an adversarial perturbation. It computes the gradient of the loss function *J* of the model concerning the image vector *x* to get the direction of pixel change and generates adversarial examples $x_{adv}$ by minimizing the probability of the true class. FGSM perturbations can be computed by minimizing either the L1, L2 or L$\infty$ norms according to the following equations[17]:

$$x^{adv} = x + \epsilon \cdot sign(\nabla_x J(\theta, x, y))$$

, where $\epsilon$ controls the perturbation magnitude and $\theta$ is the model parameters.

- Projected Gradient Descent (PGD)[23]: PGD is an improved version of FGSM by applying it multiple times with a small step size $\propto$. The adversarial examples were generated in multiple iterations by the following equation:

$$X_0^{adv} = X, X_{n+1}^{adv} = Clip_{\epsilon_p}^{X}(X_n^{adv} + \epsilon_p \cdot sign(\nabla_x J(X_n^{adv}, y))). \quad (5)$$

, where Clip$_{\epsilon_p}^{X}$( ) clips updated images to constrain it within the $\epsilon$-ball of X (i.e.,limits the change of the generated adversarial image in each iteration). The initial perturbation was a random point within the allowed $\epsilon$-ball, and the search was repeated multiple times to avoid falling into the local minimum.

- AdvFaces[21]: is a neural network model developed by Debayan et al. to generate perturbations in the silent regions of the face images without reducing the image quality. It consists of three components: a generator, a discriminator, and a face matcher. For the generator *G*, which takes an input image *x* and generates an adversarial image, $x + G(x)$, by adding an adversarial mask *G*(*x*) with minimal perturbation that is similar to the original image, using the following $L_2$ norm function:

$$L_{perturbation} = E_x[max(\epsilon, \|G(X)\|_2)] \quad (6)$$

, where $\epsilon$ represents the minimum amount of perturbation. During the training process, AdvFace used a face matcher, *F*, to supervise the training process. AdvFace minimizes the cosine similarity between the original image and the generated perturbed one using the following identity loss function:

$$L_{identity} = E_x[(F(x, x + G(x))] \quad (7)$$

As the goal of the obfuscation attack is to reject the claimed identity, the attack model uses a discriminator, *D*, that distinguishes between the probed image and the generated adversarial one by using the following GAN loss function:

$$L_{GAN} = E_x[logD(x)] + E_x[log(1 - D(x + G(x)))] \quad (8)$$

Consequently, the whole picture of the AdvFaces attack model is working based on the following objective loss function:

$$L = L_{GAN} + \lambda_i L_{identity} + \lambda_p L_{perturbation} \quad (9)$$

*Generation of Perturbation*

Figure 12 illustrates the synthesis of adversarial facial images using three distinct attack methods: FGSM, PGD, and AdvFace. Each row in the figure corresponds to adversarial images with their respective perturbation values denoted as $\epsilon$. Both FGSM and PGD rely on the gradients of the loss function and apply perturbations to every pixel in the facial image, resulting in low-quality adversarial images. In contrast, AdvFaces[21] autonomously learns to perturb specific regions of the face, such as the eyes, nose, and mouth, which are generally considered non-distracting or "silent." Consequently, AdvFaces produces higher-quality adversarial images than those generated by the FGSM and PGD attack methods.

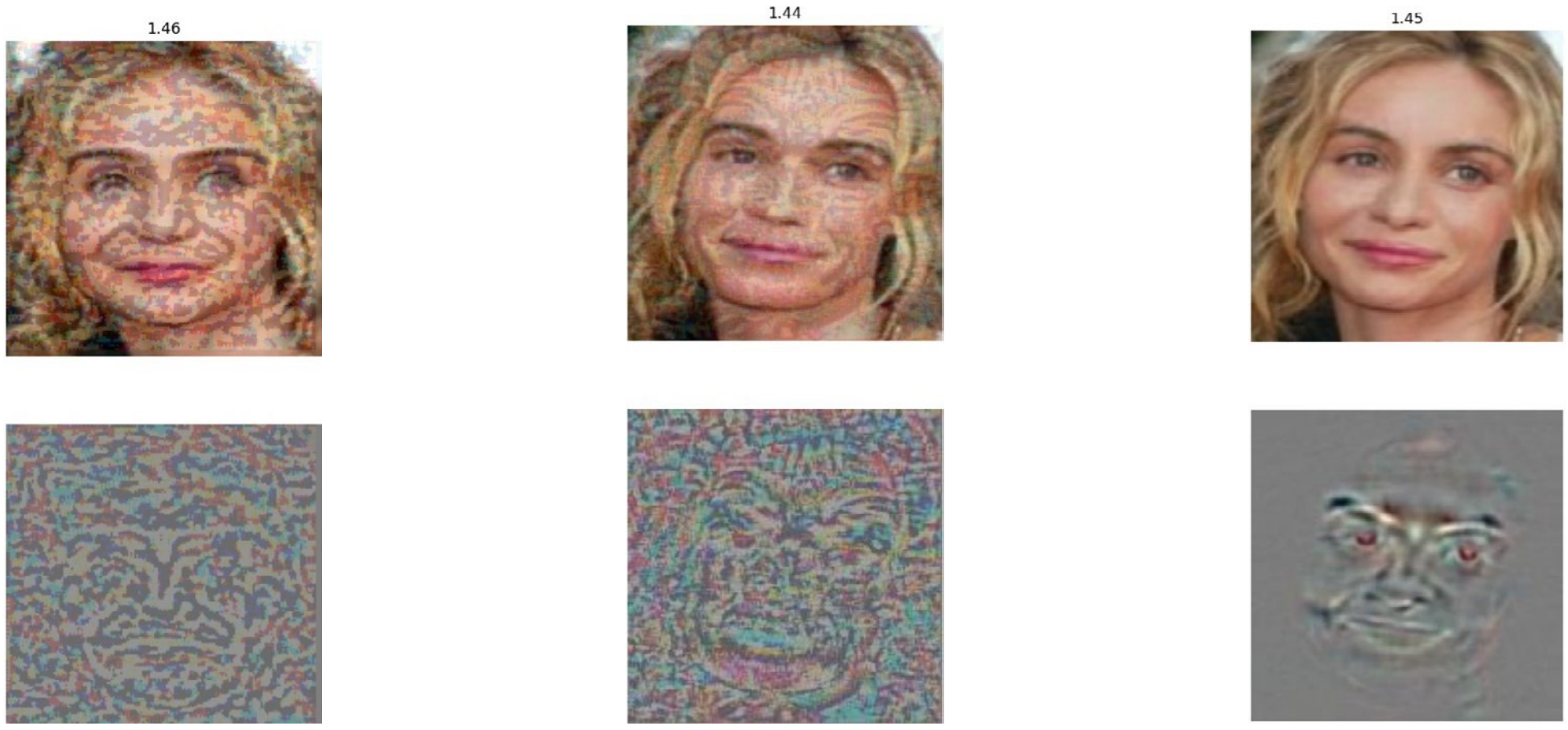

(a) FGSM [? ] (b) PGD [? ] (c) AdvFaces [? ]

**Fig. 12**. (Upper Row) Adversarial face images synthesized via three attacks: FGSM, PGD, and AdvFace. (Lower Row) Corresponding adversarial perturbations. FaceNet Euclidean distance scores between the adversarial and unaltered gallery images (not shown). A score above 1.1 indicates dissimilar subjects. (Image Credit:[90]).

## Adversarial attacks on large vision-language models (VLMs)

As face verification systems increasingly integrate vision-language models (VLMs), they have become significantly more powerful due to their ability to jointly interpret visual and textual cues. Yet, this integration has inadvertently broadened their vulnerability to sophisticated adversarial attacks. Unlike traditional face verification methods, VLM-based systems process multimodal information, which not only enriches their performance but also creates new attack surfaces that adversaries can exploit. Understanding these novel threats is essential for both researchers and practitioners to build more trustworthy verification systems.

For instance, AnyAttack[91] revealed that adversarial images crafted through self-supervised methods can deceive VLMs even without predefined labels or textual prompts. This approach, which leverages large-scale multimodal datasets, has demonstrated remarkable effectiveness and transferability across numerous commercial models. Such findings demonstrate that attackers no longer need intimate knowledge of the target systems; instead, they can exploit common embedding structures to launch potent impersonation attacks. This exposes an urgent need for defenses focusing not only on detecting adversarial patterns but also on fortifying the multimodal embeddings themselves. In addition to digital attacks, the physical world also presents surprising avenues for adversaries. The ProjAttacker method[92] exemplifies how subtle, dynamic projections of adversarial patterns onto faces can deceive verification systems, circumventing traditional defenses such as liveness detection. This attack notably shifts the adversarial paradigm from physically intrusive methods (such as masks or makeup) to subtle, environmental manipulations. Thus, attackers have become less conspicuous, prompting the need for systems that recognize subtle environmental tampering rather than overt spoofing methods.

Further complicating matters, adversaries have begun to exploit generative AI in novel ways. Adv-CPG[93], for instance, incorporates adversarial perturbations directly into generative portrait models. By embedding identity-masking mechanisms at the point of image generation, adversaries can effectively block unauthorized facial recognition, thus undermining the reliability of verification systems. This proactive integration of adversarial intent into generative processes marks a significant shift, highlighting how the boundary between data creation and attack execution is becoming increasingly blurred.

Moreover, new strategies like the DPA Attack[82] illustrate the ingenuity of adversaries who now diversify surrogate models, rather than inputs alone, to enhance attack transferability across unknown systems. Simultaneously, methods such as Cross-Modal Adversarial Patches[94] exploit the very interactions between image and text, strategically placing patches that disrupt multimodal associations while remaining visually plausible. Such developments underscore that adversaries are effectively leveraging the strengths of multimodal models against them, emphasizing that security solutions must now anticipate threats at the intersection of different modalities rather than within individual modalities alone.

Recent studies have extended adversarial research to transformer-based and multimodal face-verification systems. Kong et al.[95] demonstrates that multimodal architectures, which combine visual and auxiliary sensor inputs, can still be vulnerable to adversarial perturbations targeting individual modalities, exposing weaknesses in cross-modal fusion. Cai et al.[96] further show that Vision Transformers, a core component in many large vision-language models, can be adapted for face anti-spoofing but remain susceptible to adversarial triggers due to the high-dimensional token interactions. Complementing these findings, Digital and Physical Face Attacks[97] provides a systematic review of recent digital and real-world attack strategies, highlighting the emerging risks posed by advanced transformer- and multimodal-based face-verification pipelines.

| Defense Strategy | Authors | Method | Datasets | Attacks |
|---|---|---|---|---|
| Robustness | Kurakin et al.[25] | Adversarial training with FGSM | ImageNet[101] | FGSM[17] |
| | Jang et al.[26] | Training with adversarial examples | MNIST[105], CIFAR-10[106] | FGSM[17], C&W[80], PGD[23] |
| Transformations | Guo et al.[35] | Quilting, TVM, cropping, rescaling | ImageNet[101] | DeepFool[78], FGSM[17], C&W[80], I-FGSM[25] |
| | Shaham et al.[34] | PCA, wavelet, JPEG compression | NIPS 2017 competition | FGSM[17], C&W[80], I-FGSM[25] |
| Detection | Gong et al.[29] | Binary CNN | MNIST[105], CIFAR-10[106], SVHN | FGSM[17], TGSM[107], JSMA[79] |
| | Massoli et al.[27] | MLP/LSTM on AFR filters | VGGFace2[62] | FGSM[17], C&W[80], BIM[107] |
| | Goel et al.[28] | Adaptive noise detection | Yale Face[108] | DeepFool[78], FGSM[17], EAD[109] |
| | Goswami et al.[110] | SVM on AFR filters | MEDS[111], PaSC[112], MBGC[113] | EAD[109] |
| | Agarwal et al.[30] | PCA + SVM | PaSC[112], MEDS[111], Multi-PIE[114] | Universal Perturbation[16], Fast Feature Fool[115] |
| | Awany et al.[116] | Detection framework | CASIA-WebFace[59], LFW[11] | FGSM[17], PGD[23], AdvFaces[21] |
| Purification | Debayan et al.[36] | Generator + detector + purifier | CASIA-WebFace[59], LFW[11], CelebA[117], FFHQ[118] | FGSM[17], PGD[23], DeepFool[78], AdvFaces[21], GFLM[85], Semantic [119] |

**Table 4**. Common methods for each defense strategy in the literature.

Collectively, these insights highlight a critical shift in the landscape of adversarial attacks targeting VLM-based face verification systems. As attacks become increasingly subtle, sophisticated, and multimodal, defenders face the growing challenge of maintaining user trust and system reliability.

## Defense

Face verification systems have achieved widespread adoption, ubiquitously integrated into our smartphones. These systems facilitate various functions, including device unlocking, financial transactions, and access to premium content stored on the device. In this context, the robustness of face verification systems has emerged as a critical consideration to ensure their reliability. The failure to detect adversarial smartphone attacks poses a significant risk to confidential information, encompassing emails, bank records, social media content, and personal photos. Consequently, the presence of such adversarial examples has spurred research efforts among academic and industry groups and social media platforms to develop generalizable defenses against ever-evolving adversarial attacks. As a result, the urgency of countering face attacks has intensified, driven by mounting concerns regarding user privacy. The failure to detect such attacks poses a substantial security threat, particularly given the widespread use of face verification systems in contexts such as border control. Despite the remarkable performance exhibited by face verification systems, attributed to advances in deep learning and the availability of extensive datasets, these systems remain susceptible to the increasing menace of adversarial attacks, as indicated by various studies[13,17,21,23,98]. Attackers invest significant time and effort into manipulating faces through physical[13,98] and digital[14] means, with the objective of evading face verification systems. It has been demonstrated that these systems are vulnerable to adversarial attacks stemming from perturbations introduced to the images under scrutiny[20,21,85]. Notably, even when such perturbations are imperceptible to the human eye, they can undermine the performance of face verification. In the literature focusing on defense strategies against adversarial examples, these strategies primarily fall within two main categories: robust optimization and pre-processing techniques, as illustrated in Fig. 3. Robust optimization, a widely employed defense approach, involves altering the training procedures or architectures of neural networks to enhance their resistance to adversarial perturbations[17,23–26]. While these algorithms offer some protection against specific attack methods, they remain susceptible to other adversarial mechanisms. It is essential to note that adversarial training, a component of robust optimization, requires more time and computational resources than training models solely on clean images, as it necessitates additional computations for generating online adversarial examples . Conversely, pre-processing strategies maintain the core training procedures and network architectures unchanged and instead focus on identifying, eliminating, or purifying adversarial elements. For instance, in the context of detecting adversarial examples, this involves training a binary classifier to distinguish between genuine and adversarial instances[27–33]. In the case of removing adversarial noise[34,35], the objective is to eliminate adversarial perturbations by applying preprocessing transformations to input data before feeding it to target models. Conversely, in the context of purification, the perturbations are exclusively removed from input images containing adversarial elements[36], preventing the inadvertent alteration of genuine images and the associated high false rejection rates. Securing resilient face verification systems against adversarial examples represents a complex and ongoing challenge. A multitude of adversarial defense mechanisms have been employed to safeguard FV systems from such threats. The current body of academic literature concerning defense strategies can be categorized into two primary domains: robust optimization and preprocessing. This survey offers an overview of prior research pertinent to our investigations in adversarial defenses, with a particular emphasis on preprocessing techniques. Specifically, our research centers on perturbation removal and detection strategies. Table 4 lists common benchmark methods for each defense strategy in the literature.

## Perturbation removing

This type of defense aims to remove adversarial perturbations by applying transformations, such as preprocessing the input data and then sending these inputs to the target models. As Guo et al.[35] applied image transformations, such as total variance minimization[99], image quilting[100], image cropping and rescaling, and bit-depth reduction to smooth input images. Applied on ImageNet[101] dataset and shown that these defenses can be surprisingly effective against existing three attacks, they are as follows, (i) countering the (iterative) fast gradient sign method[25], (ii) Deepfool[78], and (iii) Carlini-Wagner attack[80] in particular, when the convolutional network is trained on the images on which these transformations are performed. Dziugaite et al.[102] and Das et al.[103] suggested applying JPEG compression to insert images before feeding them over the network. Hendrycks et al.[72]; and Li et al.[31], proposed defense methods based on principal component analysis (PCA). Liu et al. proposed the DNN-favorable JPEG compression, namely "feature distillation," by redesigning the standard JPEG compression algorithm to maximize the defense efficiency while assuring the DNN testing accuracy[104]. As a result of the good performance of these methods on an ImageNet[101] dataset in[35].

## Perturbation detection

Another strategic direction that defends against adversarial attacks on the FV system is detecting adversarial examples. Adversarial detection techniques have recently gained attention within the scientific community, and many adversarial detection mechanisms are deployed as a preprocessing step. The attacks addressed in this study[120–122] were initially suggested in object recognition and often fail to detect attacks in a feature extraction network setting, such as in face verification. Therefore, prevalent detectors against hostile faces have only been effective in a highly restricted environment where the number of people is limited and constant during training and testing[27,28,30]. Defending against adversarial attacks by detection involves creating robust systems that consist of a weak model and a detection system that indicates the occurrence of attacks. Detecting subsystems are often implemented as binary detectors that discriminate between authentic and adversarial inputs. For instance, Gong et al.[29] proposed to train an additional binary classifier that decides whether the input image is pure or adversarial. Grosse et al.[123] adopted statistical tests in pixel space to prove that adversarial images could be distinguished and suggested introducing the "adversarial" category in the original category trained according to the model. Similarly, Metzen et al.[32] proposed a detection subnetwork based on intermediate representations generated by the model at the time of inference. However, it has been shown that many detection schemes can be bypassed[120]. Guo et al.[124] propose CCA-UD, a universal backdoor-detection framework that first partitions the training data with density-based clustering and then applies a centroid-shift test. Representative features of each cluster are superimposed on benign images; a cluster is deemed poisoned when these composites consistently induce misclassification. Because the method exploits a trigger-agnostic property—general misclassification under feature overlay—it remains effective across clean-label and dirty-label attacks and with global, local, sample-specific, and source-specific triggers, achieving superior detection rates compared with prior defences.

## Adversarial training for CNN-based face verification

Conventional CNN-based face verification models, including ArcFace[8] and CosFace[125], have seen notable improvements in robustness via adversarial fine-tuning using techniques such as Projected Gradient Descent (PGD) and Fast Gradient Sign Method (FGSM)[17,23] . However, integrating adversarial perturbations directly into training has often induced a noticeable trade-off, enhancing robustness but simultaneously diminishing accuracy on clean images due to distortions in the model's embedding space. To mitigate these negative side effects, regularization strategies like TRADES[126] have been explored to strike a balance, aiming to preserve discriminative capabilities critical for facial recognition tasks. An alternative approach involves auxiliary defense mechanisms, exemplified by FaceGuard[36], which deploys a self-supervised purification and detection module without modifying the verification model itself. Such modular defenses have demonstrated high efficacy, achieving a high detection accuracy of adversarial perturbations and maintaining original accuracy for legitimate users. Nonetheless, while auxiliary modules show promise in isolating adversarial perturbations, their effectiveness relies heavily on accurately modeling diverse attack patterns, highlighting an ongoing challenge of overfitting to specific adversarial strategies and underscoring the need for continually adaptive defenses. Recent extensions, such as Adversarial Weight Perturbation and Feature-Denoising FaceNet, further improve robustness while reducing the drop in clean accuracy, indicating a move toward parameter-efficient and embedding-aware training strategies.

## Adversarial training in large vision-language models (VLMs)

Adversarial training in Vision–Language Models (VLMs) presents unique challenges due to their joint multimodal embedding spaces[127]. Early supervised adversarial fine-tuning of CLIP-like models improved robustness but often degraded zero-shot generalization when trained on fixed-label datasets[128].

Recent approaches focus on maintaining robustness while preserving generalization. Robust CLIP[129] adversarially fine-tunes the vision encoder while freezing the text encoder, leading to significant robustness improvements without compromising zero-shot capabilities. Anchor-RFT[130] constrains fine-tuning to anchor points in the joint vision–text space, which helps preserve out-of-distribution accuracy while enhancing robustness against PGD-style and Auto-Attack perturbations. Hyper-AT[131] employs a hyper-network to generate adversarial perturbations dynamically during training, reducing computational overhead while improving robust accuracy.

Beyond adversarial training, multimodal approaches have emerged as promising solutions to enhance face-verification robustness. Kong et al.[97] proposed Echo-FAS, an acoustic-based face anti-spoofing system that relies only on a speaker and microphone for liveness detection. Although Echo-FAS does not directly employ adversarial training, it illustrates how incorporating auxiliary modalities can mitigate spoofing attacks and can complement adversarially trained multimodal architectures such as M3FA[95]. Similarly, methods that detect manipulated facial regions using combined semantic and noise-level features, such as the framework by Kong et al.[132], provide complementary defences that can be integrated with transformer-based models to improve resilience to face forgeries.

Parameter-efficient strategies, including adversarial prompt tuning[133] and low-rank adaptation[131], further support robust fine-tuning by maintaining clean accuracy and zero-shot performance while requiring minimal parameter updates. Transformer-based models such as S-adapter[96] and multimodal fusion systems like M3FAS demonstrate the benefits of combining adversarial training with auxiliary defences and multiple modalities. Collectively, these developments reflect a clear trend toward modular, scalable, and attack-agnostic approaches that enhance the robustness of VLM-based face-verification systems while mitigating the trade-off with generalization ability.

## Conclusion

In conclusion, this survey has examined recent developments in face verification systems. While advancements in recent years have enhanced accuracy and efficiency, particularly through the integration of deep learning, the path forward requires a keen awareness of the human element. Challenges surrounding privacy, ethical considerations, and the persistent issue of bias in datasets and algorithms demand our continued attention.

Future research must strive for greater technical sophistication in handling variations like pose, illumination, and aging, and prioritize the development of systems that are fair, transparent, and respectful of individual rights. By focusing on creating more diverse and representative datasets, establishing robust ethical guidelines, and ensuring accountability in deployment, we can harness the power of face verification technology for societal good while safeguarding fundamental human values.

## Data availability

The datasets used and/or analysed during the current study are available from the corresponding author on reasonable request.

## Acknowledgements
We would like to thank our colleagues for their feedback on the earlier version of this survey.

## Author contributions
All authors contributed to the study conception and design. Sohair Kilany: Conceptualization, methodology, data collection, investigation, visualization, writing original draft, review and editing. Ahmed Mahfouz: Conceptualization, methodology, investigation, validation, supervision, writing review an editing.

## Funding
Open access funding provided by The Science, Technology & Innovation Funding Authority (STDF) in cooperation with The Egyptian Knowledge Bank (EKB).

## Declarations

## Competing interests
The authors declare no competing interests.

## Consent for publication

The article was submitted with the consent of all authors and institutions for publication.

## Additional information

**Correspondence** and requests for materials should be addressed to S.K. or A.M.

**Reprints and permissions information** is available at www.nature.com/reprints.

**Publisher's note** Springer Nature remains neutral with regard to jurisdictional claims in published maps and institutional affiliations.